%% file: ijcai25.tex
\documentclass{article}
\usepackage{ijcai25}

\usepackage[table]{xcolor}

\usepackage{times}
\usepackage{soul}
\usepackage{url}
\usepackage[breaklinks,colorlinks,citecolor=teal,urlcolor=teal]{hyperref}
\usepackage[utf8]{inputenc}
\usepackage[small]{caption}
\usepackage{graphicx}
\usepackage{amsmath}
\usepackage{amsthm}
\usepackage{booktabs}
\usepackage{algorithm}
\usepackage{algorithmic}
\usepackage[switch]{lineno}
\usepackage{amssymb,mathtools}
\usepackage{cleveref}
\usepackage{utfsym}
\usepackage{multirow}

\usepackage{enumitem}
\usepackage{pifont}

\title{Toward Robust Non-Transferable Learning: A Survey and Benchmark}

\author{
  \vspace{-5mm}
Ziming Hong\thanks{}
\and 
Yongli Xiang$^*$
\And
Tongliang Liu$^{\dagger}$\\
\affiliations
Sydney AI Centre, The University of Sydney\\
\emails
hoongzm@gmail.com,
yxia0023@uni.sydney.edu.au,
tongliang.liu@sydney.edu.au
}

\begin{document}

\twocolumn[{%
\renewcommand\twocolumn[1][]{#1}%
\maketitle
\begin{center}
  \vspace{-15mm}
  \centering\includegraphics[width=\linewidth]{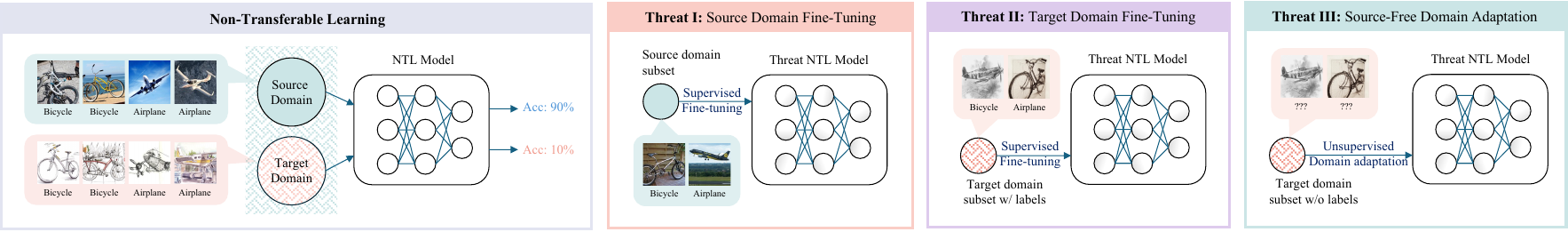}\\
  \vspace{2mm}
  \centering\includegraphics[width=1.0\linewidth]{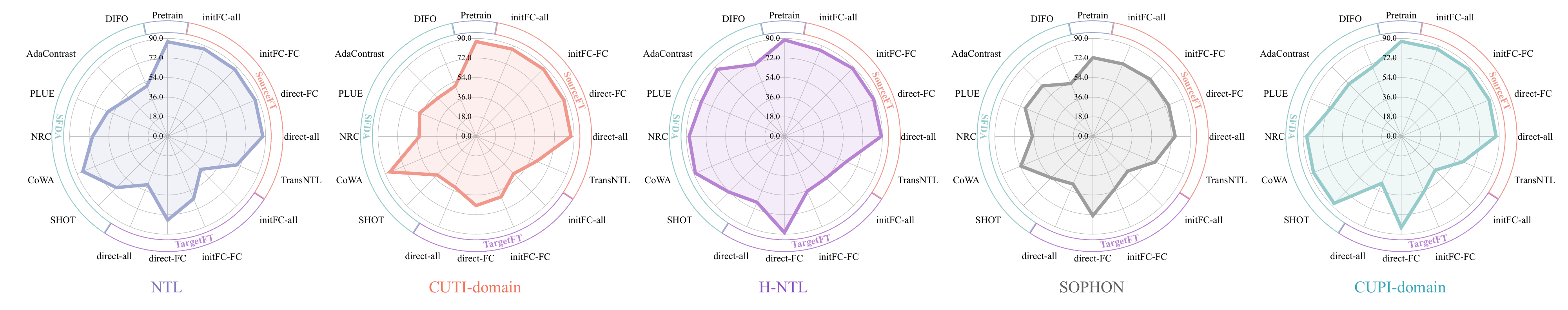}
  \vspace{-7mm}
  \captionof{figure}{We systematically review non-transferable learning (NTL) and introduce \texttt{NTLBench}, an unified framework for benchmarking NTL. This figure compares 5 methods (\textcolor[HTML]{7884AC}{NTL}, \textcolor[HTML]{7884AC}{CUTI-domain}, \textcolor[HTML]{7884AC}{H-NTL}, \textcolor[HTML]{7884AC}{SOPHON}, \textcolor[HTML]{7884AC}{CUPI-domain}) on CIFAR \& STL with VGG-13, evaluating pre-training performance and robustness
  against 5 \textcolor[HTML]{E4785F}{source domain fine-tuning} attacks, 4 \textcolor[HTML]{8151BA}{target domain fine-tuning} attacks, and 6 \textcolor[HTML]{59A4B7}{source-free domain adaptation} attacks (higher value means better robustness). 
  \texttt{NTLBench} will be released soon at \url{https://github.com/tmllab/NTLBench}.
  } 
  \vspace{3mm}
\label{fig:opening}
\end{center}%
}]

\renewcommand{\thefootnote}{*}
\footnotetext{Equal contribution. $^{\dagger}$Corresponding author.}
\renewcommand{\thefootnote}{\arabic{footnote}}

\begin{abstract}
  \vspace{-0.2mm}
  Over the past decades, researchers have primarily focused on improving the generalization abilities of models, with limited attention given to regulating such generalization. However, the ability of models to generalize to unintended data (e.g., harmful or unauthorized data) can be exploited by malicious adversaries in unforeseen ways, potentially resulting in violations of model ethics. Non-transferable learning (NTL), a task aimed at reshaping the generalization abilities of deep learning models, was proposed to address these challenges. While numerous methods have been proposed in this field, a comprehensive review of existing progress and a thorough analysis of current limitations remain lacking. In this paper, we bridge this gap by presenting the first comprehensive survey on NTL and introducing \texttt{NTLBench}, the first benchmark to evaluate NTL performance and robustness within a unified framework. Specifically, we first introduce the task settings, general framework, and criteria of NTL, followed by a summary of NTL approaches. Furthermore, we emphasize the often-overlooked issue of robustness against various attacks that can destroy the non-transferable mechanism established by NTL. Experiments conducted via \texttt{NTLBench} verify the limitations of existing NTL methods in robustness. Finally, we discuss the practical applications of NTL, along with its future directions and associated challenges.
  \vspace{-0.5mm}
\end{abstract}

\input{sections/1_introduction.tex}

\input{sections/2_preliminary.tex}

\input{sections/3_immeNTL.tex}

\input{sections/benchmarking.tex}

\input{sections/4_application.tex}

\input{sections/relatedwork.tex}
\input{sections/discussion.tex}

\bibliographystyle{named}
\bibliography{ijcai25}

\end{document}

%% file: sections/1_introduction.tex
\vspace{-1mm}
\section{Introduction}\label{sec:introduction}

Throughout much of deep learning (DL) history, researchers have primarily focused on improving generalization abilities \cite{neyshabur2017exploring,liu2021towards,zhuang2020comprehensive,yuan2025instance,yang2020rethinking,foret2020sharpness}. 
With advancements in novel techniques, the availability of high-quality data, and the expansion of model sizes and computational resources, DL models have demonstrated increasingly strong generalization, extending from in-distribution to out-of-distribution (OOD) scenarios \cite{wang2022generalizing,radford2021learning,ye2022ood,kaplan2020scaling,hendrycks2021many,li2022out,huang2023winning,zou2024towards,guo2024investigating}. 
This facilitates the application of DL in complex real-world scenarios.
However, limited attention has been given to regulating models' generalization abilities, while strong-enough yet unconstrained generalization abilities may pose misuse risks. Specifically, the generalization of deep models to unintended data (e.g., unauthorized or harmful data) can be exploited by malicious adversaries in unexpected ways. 
This raises concerns regarding the regulating of powerful DL models, including issues related to model ethic \cite{li2023trustworthy,jiao2024navigating}, safety alignment \cite{ouyang2022training,huang2024harmful,ji2024beavertails,yin2025safeworld}, model privacy and intellectual property \cite{sun2023deep,wang2024training,chen2024watermark,wuresilient,wangdefense,jiang2024intellectual}, among others.

Non-transferable learning (NTL)~\cite{wang2021non}, a task aimed at reshaping the generalization abilities of DL models, was proposed to address these challenges. 
Its goal is to prevent the model's generalization to specific target domains or tasks (such as harmful \cite{rosati2024representation,huang2024harmful} or unauthorized domains \cite{wang2021non,si2024iclguard}) while preserving its normal functionality on a source domain. Although numerous NTL methods have been proposed recently~\cite{zeng2022unsupervised,wang2023model,wang2023domain,hong2024improving,peng2024map,zhou2024archlock,hong2024your,deng2024sophon,si2024iclguard,rosati2024representation,wang2024say,ding2024non,xiang2025jailbreaking}, a comprehensive summary of existing progress in this field and an thorough analysis of current limitations is still lacking.

In this paper, we bridge this gap by presenting the first comprehensive survey of NTL.  We first introduce the task settings, general framework and criteria of NTL (\Cref{sec:Preliminary}), followed by a summary of existing NTL approaches according to their strategies to implement non-transferability in two settings (\Cref{sec:immeNTL}). 
Then, we highlight the often-overlooked robustness against diverse attacks that can destroy the non-transferable mechanism established by NTL (\Cref{sec:robustness}).

In addition, we propose the first benchmark (\texttt{NTLBench}) to integrate 5 state-of-the-art (SOTA) and open-source NTL methods and 3 types of post-training attacks (15 attack methods) in a unified framework, as illustrated in \Cref{fig:opening}. Our \texttt{NTLBench} supports running NTL and attacks on 9 datasets (more than 116 domain pairs), 5 network architecture families,
providing overall at least 40,000 experimental configurations for comprehensive evaluation. 
Main results obtained from \texttt{NTLBench} verify the unsatisfactory robustness of existing NTL methods in dealing with various post-training attacks (\Cref{sec:exp}). 
Finally, we discuss applications, related work and future directions and challenges (\Cref{sec:applications,sec:related,sec:future}).

We believe that our survey and \texttt{NTLBench} can drive the development of robust NTL methods and facilitate their applications in trustworthy model deployment scenarios. 
Our major contributions are summarized as three folds:
\begin{itemize}[leftmargin=*, topsep=0pt]\setlength{\parskip}{0pt}
    \item \textbf{Comprehensive review:} We conduct a systematic review of existing NTL works, including settings, framework, criteria, approaches, and applications. We emphasize the robustness challenges of NTL from three aspects, according to the data accessibility of different attackers.
    \item \textbf{Codebase:} We propose \texttt{NTLBench} to benchmark 5 SOTA and open-source NTL methods, covering standard assessments (5 networks and 9 datasets) and examining robustness against 15 attacks from 3 attack settings. 
    \item \textbf{Evaluation and analysis:} We use our \texttt{NTLBench} to fairly evaluate 5 SOTA NTL methods, covering the performance and robustness against diversity attacks. Our results identify the limitation of existing NTL methods in dealing with complex datasets and diverse attacks.
\end{itemize}

\begin{figure}[t!]
    \centering
    \includegraphics[width=\linewidth]{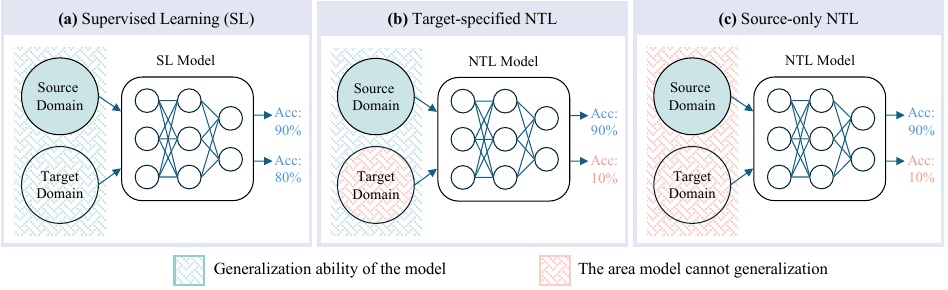}
    \vspace{-6mm}
    \caption{Comparison of (a) supervised learning (SL), (b) target-specified non-transferable learning (NTL), and (c) source-only NTL.}
    \vspace{-2mm}
    \label{fig:NTL}
\end{figure}

%% file: sections/2_preliminary.tex
\section{Preliminary}
\label{sec:Preliminary}

\subsection{Problem Setup}

In NTL, we generally consider a source domain and a target domain, where we want to keep the performance on the source domain (similar to supervised learning (SL) performance) and degrade performance on the target domain, thus implementing the resistance of generalization from the source domain to the target domain.

According to whether the target domain is known in the training stage, NTL could be divided into two settings~\cite{wang2021non}: \textbf{(i)} \textit{target-speciﬁed NTL}, which assumes the target domain is known and aims to restrict the model generalization toward the pre-known target domain, and \textbf{(ii)} \textit{source-only NTL}, which assumes the target domain is unknown and aims to restrict the generalization toward all other domains except the source domain. The comparison between SL and the two NTLs is shown in~\Cref{fig:NTL}.

\subsection{General Framework of NTL}
\label{sec:definition}

We use a classification task for illustration, as most existing NTL methods aim at image classification tasks. 
Let $\mathcal{D}_s=\{(\boldsymbol{x}_i,y_i)\}_{i=1}^{N_s}$ and $\mathcal{D}_t=\{(\boldsymbol{x}_i,y_i)\}_{i=1}^{N_t}$ represent the source domain and the target domain, respectively. Note that we here assume $\mathcal{D}_s$ and $\mathcal{D}_t$ share the same label space, as considered in \cite{wang2021non}.
Considering a neural network $f_{\theta}$ with parameters $\theta$, NTL aims to train the $f_{\theta}$ to maximize the risk on the target domain $\mathcal{D}_t$ and simultaneously minimize the risk on the source domain $\mathcal{D}_s$. 
To reach this goal, a basic NTL framework is to impose a regularization term on the SL to maximize the target domain error:
\vspace{-1.5mm}
\begin{equation}
    \begin{aligned}
 {\underset{\theta}{\min}}\ \Big\{ \mathcal{L}_{\text{ntl}} := & \underbrace{\mathbb{E}_{(\boldsymbol{x}_s,y_s)\sim \mathcal{D}_s}\left[ \mathcal{L}_{\text{src}}(f_{\theta}(\boldsymbol{x}_s), y_s)\right]}_{\mathcal{T}_{\text{src}}} \\
 \noalign{\vskip -5pt}
 - & \lambda\ \underbrace{\mathbb{E}_{(\boldsymbol{x}_t,y_t)\sim \mathcal{D}_t}\left[ \mathcal{L}_{\text{tgt}}(f_{\theta}(\boldsymbol{x}_t), y_t)\right]}_{\mathcal{T}_{\text{tgt}}} \Big\}, \\
    \noalign{\vskip -3pt}
    \end{aligned}
    \label{eq:ntlframe}
\end{equation}
where $\lambda$ is a trade-off weight, $\mathcal{L}_{\text{src}}$ and $\mathcal{L}_{\text{tgt}}$ represent the loss function (e.g., Kullback-Leible divergence) for the source and target domain, respectively. 
The learning objective contains two tasks: (i) a source domain learning task $\mathcal{T}_{\text{src}}$ to maintain the source domain performance, and (ii) a non-transferable task $\mathcal{T}_{\text{tgt}}$ to degrade the target domain performance.

Existing works generally can be seen as variants to \Cref{eq:ntlframe}, where they may focus on different fields (modal, task), data assumptions (label space, target supervision, source data dependent), and use different approaches to conduct $\mathcal{T}_{\text{tgt}}$. The statistics of these aspects considered in existing works are shown in \Cref{tab:summary}. More details for each NTL approach are illustrated in \Cref{sec:immeNTL}.

\newcommand{\twoline}[2]{\begin{tabular}[c]{@{}c@{}}#1\\#2\end{tabular}}

\newcommand{\true}{\textcolor{teal}{\usym{2714}}}
\newcommand{\false}{\textcolor{purple}{\usym{2717}}}

\begin{table*}
    \tiny
    \centering
    \begin{tabular}{@{\hspace{4pt}}c@{\hspace{4pt}}|@{\hspace{4pt}}c@{\hspace{4pt}}|@{\hspace{4pt}}c@{\hspace{4pt}}c|c@{\hspace{6pt}}c@{\hspace{6pt}}c|c@{\hspace{4pt}}c|c@{\hspace{4pt}}c@{\hspace{4pt}}}
    \toprule
    
    \multirow{2.5}{*}{\textbf{Method}}
    &
    \multirow{2.5}{*}{\textbf{Venue}}
    &
    \multicolumn{2}{c|}{\textbf{Field} \textbf{\ding{172}}}
    &
    \multicolumn{3}{c|}{\textbf{Data} \textbf{\ding{173}}}
    &
    \multicolumn{2}{c|}{\textbf{Non-Transferable Approach} \textbf{\ding{174}}}
    &
    \multicolumn{2}{c}{\textbf{Robustness} \textbf{\ding{175}}}
    \\\cmidrule{3-11}
    &
    & \textbf{Modal} & \textbf{Task} & \textbf{Label Space} & \textbf{Target Data} & \textbf{Source Data} & \textbf{Feature Space} & \textbf{Output Space} & \textbf{Source} & \textbf{Target} \\
    \midrule
    \midrule
    
    NTL \cite{wang2021non}
    & ICLR'22
    & CV & CLS & Close-Set & Labeled & 
    Dependent
    & $\max \text{MMD}(\Phi(\boldsymbol{x}_s), \Phi(\boldsymbol{x}_t))$  
    & $\max \text{KL}(f(\boldsymbol{x}_t), y_t)$
    & \true & \false \\\cmidrule{1-11}
    
    UNTL \cite{zeng2022unsupervised}
    & EMNLP'22
    & NLP & CLS & Close-Set & Unlabeled & 
    Dependent
    & \begin{tabular}[c]{@{}l@{}}$\max \text{MMD}(\Phi(\boldsymbol{x}_s), \Phi(\boldsymbol{x}_t))$\\ $ + \min \text{CE}(\Omega_d(\Phi(\boldsymbol{x})),y_d)$\end{tabular}
    & --- & \false & \false \\\cmidrule{1-11}
    
    CUTI-domain \cite{wang2023model}
    & CVPR'23
    & CV & CLS & Close-Set & Labeled & 
    Dependent
    & --- 
    & $\max \text{KL}(f(\boldsymbol{x}_t), y_t)$
    & \true & \false \\\cmidrule{1-11}
    
    DSO \cite{wang2023domain}
    & ICCV'23
    & CV & CLS & Close-Set & Unlabeled & 
    Dependent
    & --- 
    & $\min \text{KL}(f(\boldsymbol{x}_t), y_s+1)$
    & \false & \false\\\cmidrule{1-11}
    
    H-NTL \cite{hong2024improving}
    & ICLR'24
    & CV & CLS & Close-Set & Labeled & 
    Dependent
    & --- 
    & $\min \text{KL}(f(\boldsymbol{x}_t), f_{\text{sty}}(\boldsymbol{x}_t))$ 
    & \false & \false \\\cmidrule{1-11}
    
    ArchLock \cite{zhou2024archlock}
    & ICLR'24
    & CV & Cross & Open-Set & Labeled & 
    Dependent
    & --- 
    & $\max \text{CE}(f(\boldsymbol{x}_t), y_t)$ 
    & \false & \true \\\cmidrule{1-11}

    TransNTL \cite{hong2024your}
    & CVPR'24
    & CV & CLS & Close-Set & Labeled & 
    Dependent
    & --- 
    & --- 
    & \true & \false \\\cmidrule{1-11}
    
    MAP \cite{peng2024map}
    & CVPR'24
    & CV & CLS & Close-Set & Unlabeled & 
    Free
    & --- 
    & $\max \text{KL}(f(\boldsymbol{x}_t), \hat{y}_t)$
    & \false & \false \\\cmidrule{1-11}

    \multirow{2.6}{*}{SOPHON \cite{deng2024sophon}}
    & \multirow{2.6}{*}{
        \begin{tabular}[c]{@{}c@{}}    
            IEEE\\S\&P'24
        \end{tabular} 
        }
    & CV & CLS
    & Open-Set & Labeled & 
    Dependent
    & ---
    & \begin{tabular}[c]{@{}c@{}}
        $\min \text{CE}(f(\boldsymbol{x}_t),1-y_t)$
        \\
        or $\min \text{KL}(f(\boldsymbol{x}_t),\mathcal{U})$
    \end{tabular} 
    & \multirow{2.6}{*}{\false} & \multirow{2.6}{*}{\true}
    \\\cmidrule{3-9}
    &
    & CV & GEN
    & Open-Set & Labeled & 
    Dependent
    & --- 
    & $\min \text{MSE}(f(\boldsymbol{x}_t),\boldsymbol{0})$
    & &
    \\\cmidrule{1-11}
    
    CUPI-domain \cite{wang2024say}
    & TPAMI'24
    & CV & CLS & Close-Set & Labeled & 
    Dependent
    & --- 
    & $\max \text{KL}(f(\boldsymbol{x}_t), y_t)$ 
    & \true & \false \\\cmidrule{1-11}
    
    NTP \cite{ding2024non}
    & ECCV'24
    & CV & CLS & Close-Set & Labeled &  
    Dependent
    & $\min \text{FDA}(\Phi(\boldsymbol{x}_t), y_t)$
    & $\max \text{KL}(f(\boldsymbol{x}_t), y_t)$ 
    & \false & \true \\
    
    \bottomrule
    \end{tabular}
    \vspace{-2mm}
    \begin{flushleft}
        \scriptsize 
        \textbf{\ding{172}} In \textbf{Field} column, ``CV'': computer vision. ``NLP'': natural language processing. ``CLS'': classification task. ``GEN'': generation task. ``Cross'': cross task. 

        \hangindent=1em
        \hangafter=1
        \textbf{\ding{173}} In \textbf{Data} column: ``Close-Set'': source and target domain share the same label space. ``Open-Set'': source and target domain have different label space. ``Labeled'': using labeled targeted data. ``Unlabeled'': do not need labeled targeted data. ``Dependent'': using source data. ``Free'': without source data. 

        \textbf{\ding{174}} In \textbf{Non-Transferable Approach}, we split the model $f$ into a feature extractor $\Phi$ and a classifier $\Omega$, i.e., $f(\boldsymbol{x})=\Omega(\Phi(\boldsymbol{x}))$. $\Omega_d$ means an additional domain classifer. $\boldsymbol{x}_s$ and $y_s$: source-domain data and label. $\boldsymbol{x}_t$ and $y_t$: target-domain data and label. $y_d$: domain label.
         $\hat{y}_t$: target-domain pesudo label predicted by the model. $f_{\text{sty}}(\cdot)$: the style mapping function trained in H-NTL~\cite{hong2024improving}. $\mathcal{U}$: uniform distribution.
        $\boldsymbol{0}$: zero vector. 
        $\text{KL}(\cdot,\cdot)$: Kullback-Leible divergence. $\text{CE}(\cdot,\cdot)$: Cross-Entropy loss. $\text{MMD}(\cdot,\cdot)$: Maximum Mean Discrepancy. $\text{MSE}(\cdot,\cdot)$: Mean Squared Error. $\text{FDA}(\cdot,\cdot)$: Fisher Discriminant Analysis (larger value indicates better feature clustering~\cite{shao2022not}).

        \textbf{\ding{175}} In \textbf{Robustness} column, \true (or \false) represent the robustness have (or haven't) been evaluated in their original paper.
    \end{flushleft}
    \vspace{-3mm}
    \caption{Summary of NTL methods according to \textbf{Field} (modal, task), \textbf{Data} (label space, target supervision, source data dependent), \textbf{Non-Transferable Approach} (feature or output space), and \textbf{Robustness} (whether source and target domain robustness have been evaluated). 
    }
    \vspace{-1mm}
    \label{tab:summary}
  \end{table*}

\subsection{NTL Criteria}
\label{sec:criteria}

\paragraph{Non-transferability performance.} 
NTL models are usually evaluated in two aspects \cite{wang2021non}:
\begin{itemize}[leftmargin=*, topsep=0pt]\setlength{\parskip}{0pt}
    \item \textit{Source domain maintainance}: Whether the NTL model is able to achieve normal performance (i.e., the same level as the SL model) on the source domain.
    \item \textit{Target domain degradation}: The extent to which the NTL model can reduce performance on the target domain.
\end{itemize}
We review how existing methods achieve both the \textit{source domain maintainance} and the \textit{target domain degradation} in \Cref{sec:immeNTL}. Speciﬁcally, we focus on the setting that the target domain is known (i.e., target-specified NTL) in \Cref{sec:target-specified} and unknown (i.e., source-only NTL) in \Cref{sec:source-only}.

\paragraph{Post-training robustness.} 
NTL models are expected to keep the non-transferability after malicious attacks,
while not all existing works consider or evaluate the comprehensive robustness of their proposed method. 
We summarize the robustness considered in existing works into the following two parts, based on which domain is accessible to attackers.
The statistics on which aspects have been evaluated for each NTL method are shown in \Cref{tab:summary} (\textbf{Robustness} column).
\begin{itemize}[leftmargin=*, topsep=0pt]\setlength{\parskip}{0pt}
    \item \textit{Robustness against source domain attack}: 
    It has been verified that fine-tuning the white-box NTL model with a small amount of source domain data is a potential risk to break non-transferability~\cite{hong2024your,wang2021non,wang2024say}. Thus, the \textit{robustness against source domain attacks} measures how well an NTL model can resist fine-tuning attacks on the source domain.
    \item \textit{Robustness against target domain attack}: If malicious attackers have access to a small amount of labeled target domain data and the white-box NTL model, they can fine-tune the NTL model to re-activate target domain performance~\cite{deng2024sophon,ding2024non}.
    The \textit{robustness against target domain attack} evaluates how well an NTL model can defend against attack from the target domain, such as fine-tuning using target domain data.
\end{itemize}
The \text{post-training robustness} of existing methods is reviewed in \Cref{sec:robustness}, where the \textit{robustness against source domain attack} is illustrated in \Cref{sec:robustness_src}, and the \textit{robustness against target domain attack} is illustrated in \Cref{sec:robustness_tgt}.

%% file: sections/3_immeNTL.tex
\section{Approaches for NTL}
\label{sec:immeNTL}

Target-specified NTL approaches contain fundamental solutions for NTL, and thus, we first review them in \Cref{sec:target-specified}. Then, in \Cref{sec:source-only}, we review how existing works implement source-only NTL in the absence of a target domain.

\subsection{Target-Specified NTL}
\label{sec:target-specified}

Briefly, in target-specified setting, the target domain is known and we aim to restrict the generalization of a deep learning model from the source domain toward the certain target domain. 
Existing methods perform target-domain regularization either on the feature space or the output space, as we summarized in \Cref{tab:summary} (\textbf{Non-Transferable Approach} column). For more details, we introduce existing strategies as follows:

\paragraph{Output space regularization.} Output-space regularizations directly manipulate the model logits on the target domain. More specifically, these operations can be categorized into \textit{untargeted regularization} and \textit{targeted regularization}. \textit{Untargeted regularization} \cite{wang2021non,wang2023model,zhou2024archlock,peng2024map} could usually be formalized as a maximizing optimization problem, where existing methods implement this regularization by maximizing the KL divergence between the model outputs and the real labels, thus disturbing the model predictions on the target domain. However, such untargeted regularizations may face convergence issues \cite{deng2024sophon}. \textit{Targeted regularization} \cite{wang2023domain,deng2024sophon} found a proxy task on the target domain (i.e., modify the labels), thus converting the maximization objective in untargeted regularization to a minimization optimization problem. 
DSO \cite{wang2023domain} transforms the correct labels to error labels without overlap (e.g., $y_{\text{err}} = y+1$) and uses error labels as the target-domain supervision. 
H-NTL \cite{hong2024improving} first disentangle the content and style factors via a variation inference framework~\cite{blei2017variational,yao2021instance,von2021self,lin2024cs,lin2025learning}, and then, they learn the NTL model by fitting the contents of the source domain and the style of the target domain. Due to the assumption that the style is approximately to be independent to the content representations, the non-transferability could be implemented.
SOPHON \cite{deng2024sophon} aims at both image classification and generation tasks. For classification, they propose to modify the cross-entropy (CE) loss to its inverse version (i.e., modify the label $y$ to $1-y$) or calculate the KL divergence between the model outputs and a uniform distribution. For generation, SOPHON proposes to use a Denial of Service (DoS) loss, i.e., let the diffusion model fit a zero matrix at each step. Compared to untargeted regularizations, targeted regularizations always have better convergence.

\paragraph{Feature space regularization.} Feature-space regularizations further reduce the similarity between feature representations from different domains, thus restricting the transferability directly on the feature space. Feature-space regularizations can also be categorized into \textit{untargeted} and \textit{targeted} strategies, depending on whether they directly enlarge the distribution gap through a maximization objective or convert it to a minimization problem by finding a proxy target. For \textit{untargeted regularization}, existing methods \cite{wang2021non,zeng2022unsupervised} propose to maximize the maximum mean discrepancy (MMD) loss between the feature representations from different domains, where MMD measures the distribution discrepancy. 
For \textit{targeted regularization}, UNTL \cite{zeng2022unsupervised} proposes to build an auxiliary domain classifier with feature representations from different domains as inputs. By minimizing the domain-classification loss, the domain classifier could help the NTL model learn domain-distinct representations. 
NTP \cite{ding2024non} aims to minimize the Fisher Discriminant Analysis (FDA) term \cite{shao2022not} in the target domain. Specifically, a smaller FDA value indicates a reduced difference in class means and increased feature variance within each class, which is associated with poorer target domain performance.

\subsection{Source-Only NTL}
\label{sec:source-only}

Under the assumption that only source domain data is available, existing works~\cite{wang2021non,wang2023model,wang2023domain,hong2024improving} take various data augmentation methods to obtain auxiliary domains from the source domain and see them as the target domain. 
Thus, the source-only NTL problem can be solved by target-specified NTL approaches. 
These augmentation methods can be split into the following three categories:

\paragraph{Adversarial domain generation.} 
Wang \textit{et al.} \shortcite{wang2021non} use generative adversarial network (GAN) \cite{mirza2014conditional,chen2016infogan} 
to synthesize fake images from the source domain and see them as the target domain.
They train the GAN by controlling the distance and direction of the synthetic distributions to the real source domain, thus enhancing the diversity of synthetic samples and improving the degradation of any distribution with shifts to the real source domain. 
CUTI-domain \cite{wang2023model} and CUPI-domain \cite{wang2024say} add Gaussian noise to the GAN-based adaptive instance normalization (AdaIN) \cite{huang2017arbitrary} to obtain synthetic samples with random styles. They use both the synthetic samples from AdaIN and Wang \textit{et al.} \shortcite{wang2021non} as the target domain. 
MAP \cite{peng2024map} also follows the GAN framework.
 They additionally add a mutual information (MI) minimization term to enhance the variation between synthetic samples and the real source domain samples, ensuring more distinct style features.

\paragraph{Strong image augmentation.} H-NTL \cite{hong2024improving} conducts strong image augmentation \cite{sohn2020fixmatch,cubuk2020randaugment,huang2023harnessing} on real source domain data.
Strong image augmentations (e.g., blurring, sharpness, solarize) do not influence the contents but significantly change the image styles, thus imposing interventions~\cite{von2021self} on the style factors in images. Then, all augmented images are treated as the target domain for training source-only NTL.

\paragraph{Perturbation-based method.} DSO \cite{wang2023domain} proposes to minimize the worst-case risk on the uncertainty set~\cite{sagawa2019distributionally,huang2023robust,wang2023defending} over the source domain distribution, where the risk is empirically calculated through a classification loss between the model predictions and the error label.

\section{Post-Training Robustness of NTL}
\label{sec:robustness}

NTL models are expected to keep the non-transferability after malicious attacks.
However, not all existing works evaluate the robustness of their method, as we listed in \Cref{tab:summary} (the last column). 
In this section, we review the robustness of the source and target domains as considered in previous works.

\subsection{Robustness Against Source Domain Attack} 
\label{sec:robustness_src}

Earlier evaluations in \cite{wang2021non,wang2023model} show that NTL models are still resistant to state-of-art watermark removal attacks when up to 30\% source domain data are available for attack. Hong \textit{et al.} \shortcite{hong2024your} further investigate the robustness of NTL and propose TransNTL, demonstrating that non-transferability can be destroyed using less than 10\% of the source domain data. Specifically,
they find NTL \cite{wang2021non} and CUTI-domain \cite{wang2023model} inevitably result in significant generalization impairments on slightly perturbed source domains \cite{hendrycks2019benchmarking,cubuk2020randaugment}. Accordingly, they propose TransNTL to fine-tune the NTL model under an impairment repair self-distillation framework, where the source-domain predictions are used to teach the model itself how to predict on perturbed source domains. As a result, the fine-tuned model is just like a SL model without the non-transferability. Furthermore, they also propose a defense method to fix this loophole by pre-repairing the generalization impairments in perturbed source domains.
Specifically, they add a defense regularization term on existing NTL and CUT-domain training. Minimizing the defense regularization term enables the NTL model to exhibit source-domain consistent behaviors on perturbed source-domain data, thus enhancing the robustness against TransNTL attack.

\subsection{Robustness Against Target Domain Attack} 
\label{sec:robustness_tgt}

Fine-tuning the NTL models using target domain data is a more direct strategy to break the non-transferability if malicious attackers have access to some labeled target domain data. However, most existing methods \cite{wang2021non,wang2023model,zeng2022unsupervised,wang2023domain,hong2024improving,peng2024map} ignore the robustness of their methods against target-domain fine-tuning attacks. SOPHON \cite{deng2024sophon} formally proposes the problem of non-fine-tunable learning, which aims at ensuring the target-domain performance could still be poor after being fine-tuned using target domain data.
Their main idea is to involve the fine-tuning process in training stage. 
Specifically, they leverage model agnostic meta-learning (MAML) \cite{finn2017model} to simulate multiple-step fine-tuning for the current model on the target domain. Then, they add per-step risk of the target domain as the total target-domain risk. By maximizing the total target-domain risk, the robustness against target-domain attacks can be enhanced. 
ArchLock \cite{zhou2024archlock} aims to find the most non-transferable network architectures \cite{liu2018darts,real2019regularized}, where they also implicitly consider the robustness on the target domain. Speciﬁcally, they maximize the \textit{minimum risk} of an architecture on the target domain in searching the non-transferable architectures. The minimum risk is found by searching the optimal \textit{parameters} of the \textit{architecture} with the minimum task loss on the target domain.
 
However, labeled target domain data being available to malicious attackers is a strong assumption that may not always hold in practical scenarios. A more realistic scenario is that the attacker only has access to some unlabeled target domain data. Whether NTL can resist attacks driven by unlabeled target domain data has not yet been studied.

%% file: sections/benchmarking.tex
\newcommand{\cods}[1]{\textcolor[RGB]{51,68,103}{#1}}
\newcommand{\codt}[1]{\textcolor[RGB]{156,77,93}{#1}}

\newcommand{\e}[2]{{#1}}
\newcommand{\tl}[2]{\begin{tabular}[c]{@{}c@{}} \cods{#1}\\\codt{#2}\end{tabular}}
\newcommand{\tln}[2]{\begin{tabular}[c]{@{}c@{}} #1\\#2\end{tabular}}

\newcommand{\tc}[2]{\cods{#1}&\codt{#2}}

\newcommand{\tlds}[2]{\begin{tabular}[c]{@{}c@{}} \cods{#1}\\\cods{(#2)}\end{tabular}}
\newcommand{\tldt}[2]{\begin{tabular}[c]{@{}c@{}} \codt{#1}\\\codt{(#2)}\end{tabular}}

\newcommand{\bestoo}[0]{\cellcolor[HTML]{FFEEEB}}
\newcommand{\besto}[0]{\cellcolor[HTML]{E7F2F5}}
\newcommand{\bestt}[0]{\cellcolor[HTML]{faf1d8}}

\setlength{\tabcolsep}{3pt}
\setlength{\fboxsep}{0pt}

\begin{table*}[ht!]
    \tiny
    \centering
    \begin{tabular}{c|cc|cc|cc|cc|cc|cc|cc|cc|cc||cc}
      \toprule
      & 
      \multicolumn{2}{c|}{\textbf{Digits}} &
      \multicolumn{2}{c|}{\textbf{RMNIST}} &
      \multicolumn{2}{c|}{\textbf{CIFAR \& STL}} &
      \multicolumn{2}{c|}{\textbf{VisDA}} &
      \multicolumn{2}{c|}{\textbf{Office-Home}} &
      \multicolumn{2}{c|}{\textbf{DomainNet}} &
      \multicolumn{2}{c|}{\textbf{VLCS}} &
      \multicolumn{2}{c|}{\textbf{PCAS}} &
      \multicolumn{2}{c||}{\textbf{TerraInc}} &
      \multicolumn{2}{c}{\textbf{Avg.}} 
      \\

      \cmidrule(lr){2-21}  

      &
      \textbf{SA} $\uparrow$ & \textbf{TA} $\downarrow$ & 
      \textbf{SA} $\uparrow$ & \textbf{TA} $\downarrow$ & 
      \textbf{SA} $\uparrow$ & \textbf{TA} $\downarrow$ & 
      \textbf{SA} $\uparrow$ & \textbf{TA} $\downarrow$ & 
      \textbf{SA} $\uparrow$ & \textbf{TA} $\downarrow$ & 
      \textbf{SA} $\uparrow$ & \textbf{TA} $\downarrow$ & 
      \textbf{SA} $\uparrow$ & \textbf{TA} $\downarrow$ & 
      \textbf{SA} $\uparrow$ & \textbf{TA} $\downarrow$ & 
      \textbf{SA} $\uparrow$ & \textbf{TA} $\downarrow$ & 
      \textbf{SA} $\uparrow$ & \textbf{TA} $\downarrow$ 
      \\
        
      \midrule
      \midrule
      SL & 
      \tc{\e{97.7}{?}}{\e{56.0}{?}} &
      \tc{\e{99.2}{?}}{\e{62.4}{?}} &
      \tc{\e{88.2}{?}}{\e{65.7}{?}} &
      \tc{\e{86.8}{?}}{\e{37.7}{?}} &
      \tc{\e{66.4}{?}}{\e{36.9}{?}} &
      \tc{\e{45.6}{?}}{\e{ 9.9}{?}} &
      \tc{\e{79.9}{?}}{\e{56.9}{?}} &
      \tc{\e{89.5}{?}}{\e{47.3}{?}} &
      \tc{\e{93.6}{?}}{\e{14.9}{?}} &
      \tc{\e{83.0}{?}}{\e{43.0}{?}} \\

      \cmidrule(lr){1-21}
      \tln{NTL}{\cite{wang2021non}} & 
      \tc{\tlds{95.6}{-2.1}}{\tldt{12.2}{-43.8}} &
      \tc{\tlds{98.7}{-0.5}}{\tldt{12.3}{-50.1}} &
      \tc{\tlds{83.9}{-4.4}}{\tldt{ 9.9}{-55.8}} &
      \tc{\tlds{82.0}{-4.8}}{\tldt{10.9}{-26.8}} &
      \tc{\tlds{64.8}{-1.6}}{\tldt{32.4}{-4.5}} &
      \tc{\tlds{ 7.6}{-38.0}}{\tldt{ 1.4}{-8.6}} &
      \tc{\tlds{78.0}{-1.9}}{\tldt{27.1}{-29.8}} &
      \tc{\tlds{85.8}{-3.7}}{\tldt{18.0}{-29.2}} &
      \tc{\tlds{90.0}{-3.6}}{\tldt{ 8.8}{-6.1}} &
      \tc{\tlds{76.3}{-6.7}}{\tldt{14.8}{-28.3}} \\

      \cmidrule(lr){2-21}
      \tln{CUTI-domain}{\cite{wang2023model}} & 
      \tc{\tlds{97.0}{-0.8}}{\tldt{ 9.5}{-46.5}} &
      \tc{\tlds{99.2}{-0.1}}{\tldt{15.5}{-46.9}} &
      \tc{\tlds{85.1}{-3.2}}{\tldt{10.7}{-55.0}} &
      \tc{\tlds{85.3}{-1.5}}{\tldt{ 8.9}{-28.8}} &
      \tc{\besto\tlds{56.7}{-9.7}}{\besto\tldt{17.8}{-19.1}} &
      \tc{\tlds{14.0}{-31.7}}{\tldt{ 2.0}{-7.9}} &
      \tc{\besto\tlds{78.3}{-1.6}}{\besto\tldt{26.7}{-30.1}} &
      \tc{\tlds{88.4}{-1.1}}{\tldt{18.3}{-28.9}} &
      \tc{\besto\tlds{87.9}{-5.7}}{\besto\tldt{ 0.8}{-14.1}} &
      \tc{\besto\tlds{76.9}{-6.1}}{\besto\tldt{12.2}{-30.8}} \\

      \cmidrule(lr){2-21}
      \tln{H-NTL}{\cite{hong2024improving}} & 
      \tc{\tlds{97.5}{-0.2}}{\tldt{ 9.6}{-46.4}} &
      \tc{\besto\tlds{99.0}{-0.2}}{\besto\tldt{10.8}{-51.5}} &
      \tc{\besto\tlds{87.2}{-1.0}}{\besto\tldt{ 9.9}{-55.8}} &
      \tc{\besto\tlds{86.5}{-0.3}}{\besto\tldt{ 8.6}{-29.0}} &
      \tc{\tlds{51.1}{-15.2}}{\tldt{17.0}{-19.8}} &
      \tc{\besto\tlds{33.3}{-12.3}}{\besto\tldt{ 2.1}{-7.8}} &
      \tc{\tlds{79.2}{-0.8}}{\tldt{42.7}{-14.2}} &
      \tc{\tlds{89.1}{-0.3}}{\tldt{22.1}{-25.1}} &
      \tc{\tlds{88.4}{-5.2}}{\tldt{14.6}{-0.2}} &
      \tc{\tlds{79.0}{-4.0}}{\tldt{15.3}{-27.8}} \\

      \cmidrule(lr){2-21}
      \tln{SOPHON}{\cite{deng2024sophon}} & 
      \tc{\tlds{95.2}{-2.5}}{\tldt{ 9.9}{-46.1}} &
      \tc{\tlds{96.6}{-2.6}}{\tldt{38.8}{-23.6}} &
      \tc{\tlds{69.5}{-18.7}}{\tldt{24.8}{-40.9}} &
      \tc{\tlds{77.3}{-9.5}}{\tldt{10.9}{-26.8}} &
      \tc{\tlds{45.9}{-20.4}}{\tldt{17.6}{-19.3}} &
      \tc{\tlds{30.1}{-15.6}}{\tldt{ 2.5}{-7.4}} &
      \tc{\tlds{79.4}{-0.6}}{\tldt{29.5}{-27.4}} &
      \tc{\tlds{86.7}{-2.8}}{\tldt{21.6}{-25.7}} &
      \tc{\tlds{88.8}{-4.8}}{\tldt{ 7.1}{-7.7}} &
      \tc{\tlds{74.4}{-8.6}}{\tldt{18.1}{-25.0}} \\

      \cmidrule(lr){2-21}
      \tln{CUPI-domain}{\cite{wang2024say}} &  
      \tc{\besto\tlds{96.7}{-1.0}}{\besto\tldt{ 8.8}{-47.2}} &
      \tc{\tlds{98.8}{-0.4}}{\tldt{21.0}{-41.3}} &
      \tc{\tlds{86.0}{-2.3}}{\tldt{11.3}{-54.4}} &
      \tc{\tlds{84.6}{-2.2}}{\tldt{ 8.2}{-29.5}} &
      \tc{\tlds{11.6}{-54.7}}{\tldt{ 2.3}{-34.6}} &
      \tc{\tlds{ 0.8}{-44.9}}{\tldt{ 0.3}{-9.7}} &
      \tc{\tlds{77.5}{-2.5}}{\tldt{29.5}{-27.4}} &
      \tc{\besto\tlds{87.8}{-1.7}}{\besto\tldt{11.5}{-35.8}} &
      \tc{\tlds{82.4}{-11.1}}{\tldt{ 1.3}{-13.6}} &
      \tc{\tlds{69.6}{-13.4}}{\tldt{10.4}{-32.6}} \\

      \bottomrule
    \end{tabular}
    \vspace{-3mm}
    \caption{Comparison of SL and 5 NTL methods on multiple datasets. We report the \cods{source-domain accuracy} (\textbf{SA}) (\%) in \cods{blue} and \codt{target-domain accuracy} (\textbf{TA}) (\%) in \codt{red}. The best results of overall performance (OA) are highlighted in \colorbox[HTML]{E7F2F5}{blue background}. The accuracy drop compared to the pre-trained model is shown in brackets. The average performance of 9 datasets are shown in the last column (\textbf{Avg.}).}
    \label{tab:tgt-spec}
    \vspace{-4mm}
  \end{table*}

\section{Benchmarking NTL}
\label{sec:exp}

The post-training robustness has not been well-evaluated in NTL, which motivates us to build a comprehensive benchmark.
In this section, we first demonstrate the framework of our \texttt{NTLBench} (\Cref{sec:ntlbench}). Then, we present main results by conducting our \texttt{NTLBench} (\Cref{sec:ntlbenchresults}), including the pretrained NTL performance on multiple datasets, and the robustness of NTL baselines against different attacks.

\subsection{\texttt{NTLBench}}
\label{sec:ntlbench}

We propose the first NTL benchmark (\texttt{NTLBench}), which contains a standard and unified training and evaluation process. \texttt{NTLBench} supports 5 SOTA NTL methods, 9 datasets (more than 116 domain pairs), 5 network architectures families, and 15 post-training attacks from 3 attack settings, providing more than 40,000 experimental configurations. 

\paragraph{Datasets.} Our \texttt{NTLBench} is compatible with: Digits (5 domains)~\cite{deng2012mnist,hull1994database,netzer2011reading,ganin2016domain,roy2018effects}, RotatedMNIST (3 domains)~\cite{ghifary2015domain}, CIFAR and STL (2 domains)~\cite{krizhevsky2009learning,coates2011analysis}, VisDA (2 domains)~\cite{peng2017visda}, Office-Home (4 domains)~\cite{venkateswara2017deep}, DomainNet (6 domains)~\cite{peng2019moment}, VLCS (4 domains)~\cite{fang2013unbiased}, PCAS (4 domains)~\cite{li2017deeper}, and TerraInc (5 domains)~\cite{beery2018recognition}. Different domains in any dataset share the same label space, but have distribution shifts, thus being suitable for evaluating NTL methods.

\paragraph{NTL baselines.}
\texttt{NTLBench} involves all open-source NTL methods: NTL~\cite{wang2021non}, CUTI-domain~\cite{wang2023model}, H-NTL~\cite{hong2024improving}, SOPHON~\cite{deng2024sophon}, CUPI-domain~\cite{wang2024say}. Besides, we also add a vanilla supervised learning (SL) as a reference.

\paragraph{Network architecture.}
The proposed \texttt{NTLBench} is compatible with multiple network architectures, including but not limited to: 
VGG~\cite{simonyan2014very}, ResNet~\cite{he2016deep}, WideResNet~\cite{zagoruyko2016wide}, ViT~\cite{dosovitskiy2020image}, SwinT~\cite{liu2021swin}.

\paragraph{Threat I: source domain fine-tuning (SourceFT).} \textit{Attacking goal}: SourceFT tries to destroy the non-transferability by fine-tuning the NTL model using a small set of source domain data. \textit{Attacking method}: \texttt{NTLBench} involves 5 methods, including four basic fine-tuning strategies\footnote{\label{initfc}initFC: re-initialize the last full-connect (FC) layer. direct: no re-initialize. all: fine-tune the whole model. FC: fine-tune last FC.}: initFC-all, initFC-FC, direct-FC, direct-all~\cite{deng2024sophon} and the special designed attack for NTL: TransNTL~\cite{hong2024your}.

\paragraph{Threat II: target domain fine-tuning (TargetFT).} \textit{Attacking goal}: TargetFT tries to directly use labeled target domain data to fine-tune the NTL model, thus recovering target domain performance. \textit{Attacking method}: \texttt{NTLBench} use 4 basic fine-tuning strategies\textsuperscript{\ref{initfc}} leveraged in~\cite{deng2024sophon} as attack methods: initFC-all, initFC-FC, direct-FC, direct-all.

\paragraph{Threat III: source-free domain adaptation (SFDA).} \textit{Attacking goal}: We introduce SFDA to test whether using unlabeled target domain data poses a threat to NTL. \textit{Attacking method}: \texttt{NTLBench} involves 6 SOTA SFDA methods: SHOT~\cite{liang2020we}, CoWA~\cite{lee2022confidence}, NRC~\cite{yang2021exploiting}, PLUE~\cite{litrico2023guiding}, AdaContrast~\cite{chen2022contrastive}, and DIFO~\cite{tang2024source}.

\paragraph{Evaluation metric.} For source domain, we use source domain accuracy (\textbf{SA}) to evaluate the performance. Higher SA means lower influence of non-transferability to the source domain utility.
For target domain, we use target domain accuracy (\textbf{TA}) to evaluate the performance. Lower TA means better performance of non-transferability.
Besides, we calculate the overall performance (denoted as \textbf{OA}) of an NTL method as: $\text{OA}=(\text{SA}+(100\%-\text{TA}))/2$, with higher OA representing better overall performance of an NTL method. These evaluation metrics are applicable for both non-transferability performance and robustness against different attacks.

\subsection{Main Results and Analysis.}
\label{sec:ntlbenchresults}

Due to the limited space, we present main results obtained from our \texttt{NTLBench}. 
We first show the key implementation details, and then we present and analyse of our results.

\paragraph{Implementation details.}
Briefly, in pre-training stage, we sequentially pair $i$-th and ($i$+1)-th domains within a dataset for training. Each domain is randomly split into 8:1:1 for training, validation, and testing. The results for each dataset are averaged across domain pairs. NTL methods and the reference SL method are pretrained by up to 50 epochs. We search suitable hyper-parameters for each method by setting 5 values around their original value and choose the best value according to the best OA on validation set. All the batch size, learning rate, and optimizer are follow their original implementations. Following the original NTL paper~\cite{wang2021non}, we use VGG-13 without batch-normalization. All input images are resize to 64$\times$64. 
In attack stage, we use 10\% amount of the training set to perform attack. All attack results we reported are run on CIFAR \& STL. Attack training is up to 50 epochs.
We run all experiments on RTX 4090 (24G).

\paragraph{Non-transferability performance.} The non-transferability performance are shown in \Cref{tab:tgt-spec}, where we compare 5 NTL methods and SL on 9 datasets. From the results, all NTL methods generally effectively degrade source-to-target generalization, leading to a significant drop in TA compared to SL. However, in more complex datasets such as Office-Home and DomainNet, existing NTL methods fail to achieve a satisfactory balance between maintaining SA and degrading TA, highlighting their limitations. From the \textbf{Avg.} column, CUTI-domain reaches the overall best performance.

\paragraph{Post-training robustness.} 
For \textbf{SourceFT} attack (\Cref{tab:atk_src}), fine-tuning each NTL model by using basic fine-tuning strategies on 10\% source domain data cannot directly recover the source-to-target generalization. However, all NTL methods are fragile when facing the TransNTL attack. For \textbf{TargetFT} attack (\Cref{tab:atk_tgt_label}), all NTL methods cannot fully resist supervised fine-tuning attack by using target domain data. In particular, fine-tuning all parameters usually results in better attack effectiveness. For \textbf{SFDA} (\Cref{tab:atk_tgt_sfda}), although the target domain data are unlabeled, advanced source-free unsupervised domain adaptation, leveraging self-supervised strategies, can still partially recover target domain performance. All these results verify the fragility of existing NTL methods.

\begin{table}[t!]
    \tiny
    \centering
    \begin{tabular}{@{\hspace{4pt}}c@{\hspace{3pt}}|c@{\hspace{2pt}}c@{\hspace{3pt}}|@{\hspace{3pt}}c@{\hspace{2pt}}c@{\hspace{3pt}}|@{\hspace{3pt}}c@{\hspace{3pt}}c@{\hspace{3pt}}|@{\hspace{3pt}}c@{\hspace{3pt}}c@{\hspace{3pt}}|@{\hspace{3pt}}c@{\hspace{3pt}}c@{\hspace{3pt}}}

    \toprule
  
      & 
      \multicolumn{2}{c|@{\hspace{3pt}}}{\textbf{NTL}} &
      \multicolumn{2}{c|@{\hspace{3pt}}}{\textbf{CUTI}} &
      \multicolumn{2}{c|@{\hspace{3pt}}}{\textbf{H-NTL}} &
      \multicolumn{2}{c|@{\hspace{3pt}}}{\textbf{SOPHON}} &
      \multicolumn{2}{@{\hspace{3pt}}c}{\textbf{CUPI}}
      \\
  
      \cmidrule(lr){2-11}
  
      &
      \textbf{SA} $\uparrow$ & \textbf{TA} $\downarrow$ & 
      \textbf{SA} $\uparrow$ & \textbf{TA} $\downarrow$ & 
      \textbf{SA} $\uparrow$ & \textbf{TA} $\downarrow$ & 
      \textbf{SA} $\uparrow$ & \textbf{TA} $\downarrow$ & 
      \textbf{SA} $\uparrow$ & \textbf{TA} $\downarrow$ 
      
      \\
      \midrule
      \midrule
      Pre-train & 
      \tc{\e{83.9}{?}}{\e{ 9.9}{?}} &
      \tc{\e{85.1}{?}}{\e{10.6}{?}} &
      \tc{\e{87.2}{?}}{\e{ 9.9}{?}} &
      \tc{\e{69.5}{?}}{\e{24.8}{?}} &
      \tc{\e{86.0}{?}}{\e{11.3}{?}} \\
      \cmidrule(lr){1-11}
  
      initFC-all & 
      \tc{\tlds{84.0}{+0.2}}{\tldt{ 9.8}{-0.1}} &
      \tc{\tlds{84.2}{-0.9}}{\tldt{10.6}{+0.0}} &
      \tc{\tlds{87.8}{+0.6}}{\tldt{16.2}{+6.3}} &
      \tc{\tlds{82.2}{+12.7}}{\tldt{38.1}{+13.3}} &
      \tc{\tlds{85.3}{-0.7}}{\tldt{11.4}{+0.1}} \\
      \cmidrule(lr){2-11}
  
      initFC-FC & 
      \tc{\tlds{84.2}{+0.3}}{\tldt{10.0}{+0.1}} &
      \tc{\tlds{85.4}{+0.3}}{\tldt{10.6}{+0.0}} &
      \tc{\tlds{87.2}{-0.1}}{\tldt{10.2}{+0.3}} &
      \tc{\tlds{71.9}{+2.4}}{\tldt{23.3}{-1.6}} &
      \tc{\tlds{85.9}{-0.1}}{\tldt{11.3}{+0.0}} \\
      \cmidrule(lr){2-11}
  
      direct-FC & 
      \tc{\tlds{84.0}{+0.2}}{\tldt{ 9.9}{+0.0}} &
      \tc{\tlds{85.2}{+0.2}}{\tldt{10.6}{+0.0}} &
      \tc{\tlds{87.3}{+0.1}}{\tldt{ 9.9}{+0.0}} &
      \tc{\tlds{74.3}{+4.8}}{\tldt{23.8}{-1.1}} &
      \tc{\tlds{86.1}{+0.1}}{\tldt{11.3}{+0.0}} \\
      \cmidrule(lr){2-11}
  
      direct-all & 
      \tc{\tlds{84.7}{+0.8}}{\tldt{ 9.8}{-0.1}} &
      \tc{\tlds{85.3}{+0.3}}{\tldt{10.9}{+0.3}} &
      \tc{\tlds{88.0}{+0.8}}{\tldt{10.1}{+0.2}} &
      \tc{\tlds{83.4}{+13.9}}{\tldt{32.2}{+7.4}} &
      \tc{\tlds{85.5}{-0.5}}{\tldt{11.3}{+0.0}} \\
      \cmidrule(lr){2-11}
  
      TransNTL & 
      \tc{\bestoo \tlds{81.7}{-2.2}}{\bestoo \tldt{44.3}{+34.4}} &
      \tc{\bestoo \tlds{81.3}{-3.8}}{\bestoo \tldt{61.0}{+50.3}} &
      \tc{\bestoo \tlds{86.3}{-1.0}}{\bestoo \tldt{63.7}{+53.8}} &
      \tc{\bestoo \tlds{83.8}{+14.3}}{\bestoo \tldt{60.1}{+35.3}} &
      \tc{\bestoo \tlds{83.1}{-2.9}}{\bestoo \tldt{60.6}{+49.3}} \\
  
      \bottomrule
    \end{tabular}
    \vspace{-3mm}
    \caption{NTL robustness against source domain fine-tuning (Source-\\FT). We show \cods{source-domain accuracy} (\textbf{SA}) (\%) and \codt{target-domain accuracy} (\textbf{TA}) (\%). The most serious threat (worst OA) to each NTL is marked as\colorbox[HTML]{fee8e4}{ red.} Accuracy drop from the pre-trained model is in ($\cdot$).}
    \label{tab:atk_src}
    \vspace{1mm}
    \begin{tabular}{@{\hspace{4pt}}c@{\hspace{3pt}}|c@{\hspace{2pt}}c@{\hspace{3pt}}|@{\hspace{3pt}}c@{\hspace{2pt}}c@{\hspace{3pt}}|@{\hspace{3pt}}c@{\hspace{3pt}}c@{\hspace{3pt}}|@{\hspace{3pt}}c@{\hspace{3pt}}c@{\hspace{3pt}}|@{\hspace{3pt}}c@{\hspace{3pt}}c@{\hspace{3pt}}}
      \toprule
    
        & 
        \multicolumn{2}{c|@{\hspace{3pt}}}{\textbf{NTL}} &
        \multicolumn{2}{c|@{\hspace{3pt}}}{\textbf{CUTI}} &
        \multicolumn{2}{c|@{\hspace{3pt}}}{\textbf{H-NTL}} &
        \multicolumn{2}{c|@{\hspace{3pt}}}{\textbf{SOPHON}} &
        \multicolumn{2}{@{\hspace{3pt}}c}{\textbf{CUPI}}
        \\
    
        \cmidrule(lr){2-11}
    
        &
        \textbf{SA} $\uparrow$ & \textbf{TA} $\downarrow$ & 
        \textbf{SA} $\uparrow$ & \textbf{TA} $\downarrow$ & 
        \textbf{SA} $\uparrow$ & \textbf{TA} $\downarrow$ & 
        \textbf{SA} $\uparrow$ & \textbf{TA} $\downarrow$ & 
        \textbf{SA} $\uparrow$ & \textbf{TA} $\downarrow$ 
        
        \\
        \midrule
        \midrule
        Pre-train & 
        \tc{\e{83.9}{?}}{\e{ 9.9}{?}} &
        \tc{\e{85.1}{?}}{\e{10.7}{?}} &
        \tc{\e{87.2}{?}}{\e{ 9.9}{?}} &
        \tc{\e{69.5}{?}}{\e{24.8}{?}} &
        \tc{\e{86.0}{?}}{\e{11.3}{?}} \\
        \cmidrule(lr){1-11}
    
        initFC-all & 
        \tc{\bestoo \tlds{23.9}{-60.0}}{\bestoo \tldt{37.8}{+27.9}} &
        \tc{\bestoo \tlds{13.3}{-71.8}}{\bestoo \tldt{15.9}{+5.3}} &
        \tc{\tlds{19.0}{-68.3}}{\tldt{10.4}{+0.5}} &
        \tc{\tlds{59.0}{-10.5}}{\tldt{68.5}{+43.7}} &
        \tc{\tlds{41.2}{-44.8}}{\tldt{53.1}{+41.8}} \\
        \cmidrule(lr){2-11}
    
        initFC-FC & 
        \tc{\tlds{33.9}{-50.0}}{\tldt{ 9.6}{-0.4}} &
        \tc{\tlds{30.2}{-54.9}}{\tldt{ 9.7}{-1.0}} &
        \tc{\tlds{19.1}{-68.1}}{\tldt{ 9.7}{-0.2}} &
        \tc{\tlds{21.6}{-48.0}}{\tldt{16.8}{-8.1}} &
        \tc{\tlds{21.8}{-64.2}}{\tldt{12.1}{+0.8}} \\
        \cmidrule(lr){2-11}
    
        direct-FC & 
        \tc{\tlds{64.2}{-19.7}}{\tldt{10.2}{+0.3}} &
        \tc{\tlds{38.0}{-47.1}}{\tldt{10.6}{-0.1}} &
        \tc{\tlds{87.1}{-0.1}}{\tldt{10.0}{+0.1}} &
        \tc{\tlds{70.5}{+1.0}}{\tldt{24.5}{-0.4}} &
        \tc{\tlds{78.6}{-7.4}}{\tldt{11.0}{-0.4}} \\
        \cmidrule(lr){2-11}
    
        direct-all & 
        \tc{\tlds{13.9}{-70.0}}{\tldt{17.6}{+7.7}} &
        \tc{\tlds{10.1}{-75.0}}{\tldt{ 8.8}{-1.9}} &
        \tc{\bestoo \tlds{84.7}{-2.5}}{\bestoo \tldt{53.3}{+43.4}} &
        \tc{\bestoo \tlds{68.0}{-1.6}}{\bestoo \tldt{72.9}{+48.1}} &
        \tc{\bestoo \tlds{51.9}{-34.1}}{\bestoo \tldt{58.4}{+47.1}} \\
    
        \bottomrule
      \end{tabular}
      \vspace{-3mm}
      \caption{NTL robustness against target domain fine-tuning (Target-\\FT).  We report \cods{source-domain accuracy} (\textbf{SA}) (\%) and \codt{target-domain accuracy} (\textbf{TA}) (\%). The most serious threat (best TA) to each NTL is marked as\colorbox[HTML]{fee8e4}{ red.} Accuracy drop from the pre-trained model is in ($\cdot$).}
      \vspace{-3mm}
      \label{tab:atk_tgt_label}
  \end{table}
  
\paragraph{More results.} 

Additional results and analysis on: various architectures, attack using different data amount, cross-domain/task, and visualizations (e.g., feature activation, t-SNE \cite{van2008visualizing}, GradCAM \cite{selvaraju2017grad}) will be released soon at our online page.

  \begin{table}[t!]
    \tiny
    \centering
    \begin{tabular}{@{\hspace{4pt}}c@{\hspace{3pt}}|c@{\hspace{2pt}}c@{\hspace{3pt}}|@{\hspace{3pt}}c@{\hspace{2pt}}c@{\hspace{3pt}}|@{\hspace{3pt}}c@{\hspace{3pt}}c@{\hspace{3pt}}|@{\hspace{3pt}}c@{\hspace{3pt}}c@{\hspace{3pt}}|@{\hspace{3pt}}c@{\hspace{3pt}}c@{\hspace{3pt}}}
    \toprule
  
      & 
      \multicolumn{2}{c|@{\hspace{3pt}}}{\textbf{NTL}} &
      \multicolumn{2}{c|@{\hspace{3pt}}}{\textbf{CUTI}} &
      \multicolumn{2}{c|@{\hspace{3pt}}}{\textbf{H-NTL}} &
      \multicolumn{2}{c|@{\hspace{3pt}}}{\textbf{SOPHON}} &
      \multicolumn{2}{@{\hspace{3pt}}c}{\textbf{CUPI}}
      \\
  
      \cmidrule(lr){2-11}
  
      &
      \textbf{SA} $\uparrow$ & \textbf{TA} $\downarrow$ & 
      \textbf{SA} $\uparrow$ & \textbf{TA} $\downarrow$ & 
      \textbf{SA} $\uparrow$ & \textbf{TA} $\downarrow$ & 
      \textbf{SA} $\uparrow$ & \textbf{TA} $\downarrow$ & 
      \textbf{SA} $\uparrow$ & \textbf{TA} $\downarrow$ 
      
      \\
      \midrule
      \midrule
      Pre-train & 
      \tc{\e{83.9}{?}}{\e{ 9.9}{?}} &
      \tc{\e{85.1}{?}}{\e{10.7}{?}} &
      \tc{\e{87.2}{?}}{\e{ 9.9}{?}} &
      \tc{\e{69.5}{?}}{\e{24.8}{?}} &
      \tc{\e{85.5}{?}}{\e{11.3}{?}} \\
      \cmidrule(lr){1-11}
  
      SHOT & 
      \tc{\tlds{63.0}{-20.9}}{\tldt{29.6}{+19.7}} &
      \tc{\tlds{35.3}{-49.8}}{\tldt{34.7}{+24.0}} &
      \tc{\tlds{86.6}{-0.6}}{\tldt{41.9}{+32.0}} &
      \tc{\bestoo \tlds{64.8}{-4.8}}{\bestoo \tldt{56.7}{+31.9}} &
      \tc{\tlds{85.8}{+0.3}}{\tldt{11.3}{+0.0}} \\
      \cmidrule(lr){2-11}
  
      CoWA & 
      \tc{\tlds{81.1}{-2.8}}{\tldt{12.4}{+2.5}} &
      \tc{\tlds{84.0}{-1.1}}{\tldt{12.7}{+2.1}} &
      \tc{\tlds{87.2}{+0.0}}{\tldt{10.1}{+0.2}} &
      \tc{\tlds{69.2}{-0.4}}{\tldt{26.1}{+1.3}} &
      \tc{\tlds{85.7}{+0.2}}{\tldt{11.3}{+0.0}} \\
      \cmidrule(lr){2-11}
  
      NRC & 
      \tc{\tlds{57.7}{-26.2}}{\tldt{19.8}{+9.9}} &
      \tc{\tlds{39.4}{-45.7}}{\tldt{35.5}{+24.8}} &
      \tc{\tlds{87.3}{+0.1}}{\tldt{12.1}{+2.2}} &
      \tc{\tlds{66.6}{-3.0}}{\tldt{55.6}{+30.8}} &
      \tc{\tlds{86.0}{+0.5}}{\tldt{12.2}{+0.9}} \\
      \cmidrule(lr){2-11}
  
      PLUE &  
      \tc{\bestoo \tlds{71.5}{-12.4}}{\bestoo \tldt{52.8}{+42.9}} &
      \tc{\bestoo \tlds{76.1}{-9.0}}{\bestoo \tldt{63.8}{+53.1}} &
      \tc{\tlds{85.5}{-1.8}}{\tldt{20.1}{+10.2}} &
      \tc{\tlds{75.5}{+6.0}}{\tldt{41.1}{+16.3}} &
      \tc{\bestoo \tlds{82.4}{-3.2}}{\bestoo \tldt{43.6}{+32.3}} \\
      \cmidrule(lr){2-11}
  
      \tln{Ada-}{Contrast} & 
      \tc{\tlds{ 9.4}{-74.5}}{\tldt{ 9.8}{-0.1}} &
      \tc{\tlds{ 9.3}{-75.8}}{\tldt{10.0}{-0.7}} &
      \tc{\tlds{86.3}{-1.0}}{\tldt{12.1}{+2.2}} &
      \tc{\tlds{64.5}{-5.1}}{\tldt{33.4}{+8.6}} &
      \tc{\tlds{47.2}{-38.3}}{\tldt{11.3}{+0.0}} \\
      \cmidrule(lr){2-11}
  
      DIFO & 
      \tc{\tlds{ 9.2}{-74.7}}{\tldt{ 9.2}{-0.7}} &
      \tc{\tlds{ 9.2}{-75.9}}{\tldt{ 9.2}{-1.5}} &
      \tc{\bestoo \tlds{85.0}{-2.2}}{\bestoo \tldt{42.1}{+32.2}} &
      \tc{\tlds{56.3}{-13.2}}{\tldt{51.3}{+26.5}} &
      \tc{\tlds{48.4}{-37.1}}{\tldt{10.4}{-1.0}} \\
  
      \bottomrule
    \end{tabular}
    \vspace{-3mm}
    \caption{NTL robustness against source-free domain adaptation (SFDA). We show \cods{source-domain accuracy} (\textbf{SA}) (\%), \codt{target-domain accuracy} (\textbf{TA}) (\%), and accuracy drop from the pre-trained model is in ($\cdot$). The most serious threat (highest TA) to each NTL is in\colorbox[HTML]{fee8e4}{ red.}}
    \vspace{-2.5mm}
    \label{tab:atk_tgt_sfda}
  \end{table}

%% file: sections/4_application.tex
\section{Applications of NTL}
\label{sec:applications}

NTL supports different applications, depending on which data are used as source and target domain. We introduce two applications in model intellectual property (IP) protection and then the application of harmful fine-tuning defense. 

\paragraph{Ownership verification (OV).} OV is a passive IP protection manner, which aims to verify the ownership of a deep learning model \cite{cheng2021mid,lederer2023identifying}. NTL solves ownership verification by triggering misclassification on data with pre-defined triggers \cite{wang2021non,chen2024mark,guo2024zeromark}. For example, when training, we add a shallow trigger (only known by the model owner) on the original dataset data and see them as the target domain, while the original data without the trigger is regarded as the source domain. Then, target-specified NTL is used to train a model. Therefore, the ownership can be verified via observing the performance difference of a model on the original data and the data with the pre-defined trigger. For SL model, the shallow trigger has minor influence on the model performance, and thus, the model shows similar performance on original data and data with triggers. In contrast, the NTL model specific to this pre-defined trigger has high performance on the original data but random-guess-like performance on data with the trigger. This provides evidence for verifying the model's ownership.
\paragraph{Applicability authorization (AA).} AA is an active IP protection approach that ensures models can only be effective on authorized data \cite{wang2021non,xu2024idea,si2024iclguard}. NTL solves AA by degrading the model generalization outside the authorized domain. Basic solution is to add a pre-defined trigger on original data (seen as source domain), and the original data without the correct triggers is regarded as the target domain. After training by NTL, the model will only perform well on authorized data (i.e., the data with the trigger). Any unauthorized data will be randomly predicted by the NTL model. Thus, AA can be achieved.

\paragraph{Safety alignment and harmful fine-tuning defense.} 
Fine-tuning large language models (LLMs) with user's own data for downstream tasks has recently become a popular online service \cite{huang2024harmful,openai2024finetune}. However, this practice raises concerns about compromising the safety alignment of LLMs \cite{qi2023fine,yang2023shadow,zhan2023removing}, as harmful data may be present in users' datasets, whether intentionally or unintentionally. To address the risks of harmful fine-tuning, various defensive solutions \cite{huang2024booster,rosati2024representation,huang2024vaccine} have been proposed to ensure that fine-tuned LLMs can effectively refuse harmful queries. Specifically, these defense methods aim to limit the transferability of LLMs from harmless queries to harmful ones, which techniques are variants of the objectives of NTL. 
Actually, all existing NTL approaches can be applied to this task by regarding the alignment data as the source domain and the harmful data as the target domain. Then, target-specified NTL can be conducted to defend agaginst harmful fine-tuning attacks.

%% file: sections/relatedwork.tex
\section{Related Works}
\label{sec:related}

\paragraph{Machine unlearning (MU).}
Both MU \cite{bourtoule2021machine,nguyen2022survey,qu2023learn,xu2023machine,zhu2024decoupling,wang2024unlearning} and NTL serve purposes in model capacity control, albeit with  differences in their applications and methodologies. 
MU primarily aims to forget specific data points
from training datasets \cite{xu2023machine,rosati2024representation} (the model behaviors are consistent to never training on the selected data points), while NTL aims at resist the generalization from the training domain to a specific target domain or task. Particularly, MU and NTL share some overlapping applications such as safety alignment of LLMs. However, MU more focus on eliminating harmful data influence (e.g., sensitive or illegal information) and the associated model capabilitie \cite{barez2025open,maini2024tofu}, while NTL more focus on preventing harmful and unauthorized fine-tuning \cite{huang2024harmful}.

\paragraph{Transfer learning (TL).} TL \cite{zhuang2020comprehensive} aims at improving model performance on a different but related target domain or task. It can be categorized into several subfields, such as unsupervised domain adaptation (UDA) \cite{long2016unsupervised,venkateswara2017deep,kang2019contrastive,liu2022deep}, source-free domain adaptation (SFDA) \cite{liang2020we,liang2021source,tang2024sourcefree,tang2024unified}, test-time adaptation (TTA) \cite{chen2022contrastive,liang2025comprehensive}, domain generalization (DG) \cite{wang2022generalizing,ye2023coping,huang2023robust}, few-shot/zero-shot learning (FSL/ZSL) \cite{xian2018zero,chen2022transzero,chen2024causal,chen2024rethinking,hou2024visual}, continual learning (CL) \cite{lopez2017gradient,zenke2017continual,wang2022improving,wang2024comprehensive}, etc. 
Specifically, domain adaptation and domain generalization are closely related to NTL, but their overall objectives are opposite to NTL. In general, TL techniques generally can be used in a reversed way to achieve NTL. Moreover, TL can also be seen as post-training attacks against NTL, as we discussed in \Cref{sec:robustness,sec:exp}. In addition, NTL can also be used as an attack for TL.

%% file: sections/discussion.tex
\section{Future Directions and Challenges}
\label{sec:future}

\paragraph{Improving robustness.} We highlight the shortcoming of NTL on post-training robustness.
Existing defense attempts (e.g., SOPHON~\cite{deng2024sophon}) require extensive resources, such as an extremely high number of training epochs, yet they may still fail to remain robust against unseen fine-tuning. This raises an open challenge: how to effectively enhance the robustness of NTL against various attacks.
\paragraph{Identifying more threat.}
There are other potential attacks that could pose risks to NTL under weaker assumptions. For example, if an attacker is unable to re-train the NTL model \cite{hu2023learning,guo2023scale}, can they still bypass the non-transferability constraints? Moreover, attackers may have access to a large amount of data from the wild \cite{chen2021learning}, distinct from both the source and target domains. Can they leverage these unseen data to break non-transferability? We believe identifying these threats can further promote the robustness of NTL.
\paragraph{Cross-modal non-transferability.} Existing NTL works primarily focus on single modality, while the cross-modal non-transferability remains an important yet underexplored challenge. A related finding in large models suggests that the safety alignment of LLMs can be compromised through visual instruction tuning~\cite{zong2024safety,liu2023visual}. 
However, a deep investigation of robust cross-modal non-transferability mechanisms remains lacking.
Advancing cross-modal non-transferability not only addresses this gap but also broadens the application scope of NTL.

\section{Conclusion}
In this paper, we conduct the first systematic review of NTL by summarizing existing approaches and highlighting the often overlooked robustness challenges. 
In addition, we propose \texttt{NTLBench} to benchmark 5 SOTA NTL methods, covering standard assessments and examining robustness against 15 attacks. 
Main results from \texttt{NTLBench} verify the limitation of existing NTLs on robustness.
We believe \texttt{NTLBench} can drive the development of robust NTL and facilitate their applications in practical scenarios.